\documentclass[runningheads]{llncs}
\usepackage{graphicx}
\usepackage{amsmath,amssymb} 
\usepackage{color}
\usepackage[width=122mm,left=12mm,paperwidth=146mm,height=193mm,top=12mm,paperheight=217mm]{geometry}

\usepackage{graphicx}
\usepackage[utf8]{inputenc}
\usepackage[font=small,labelfont=bf]{caption}
\usepackage{subfig}
\usepackage{dirtytalk}

\usepackage{siunitx}

\usepackage[breaklinks=true,bookmarks=false]{hyperref}
\newcommand{\bb}[1]{\textbf{#1}}
\usepackage{hyperref}

\hypersetup{
    colorlinks=true,
    linkcolor=black,
    citecolor=black,
    filecolor=black,
    urlcolor=black,
}

\begin{document}
\newcommand*\samethanks[1][\value{footnote}]{\footnotemark[#1]}
\pagestyle{headings}
\mainmatter

\title{Dynamic Multimodal Instance Segmentation Guided by Natural Language Queries} 

\titlerunning{Dynamic Multimodal Instance Segmentation}

\authorrunning{Edgar Margffoy-Tuay, Juan C. Pérez, Emilio Botero, Pablo Arbeláez}


\author{Edgar Margffoy-Tuay \and Juan C. Pérez \and Emilio Botero \and Pablo Arbeláez}


\institute{
	Universidad de los Andes, Colombia\\
	\email{ \{ea.margffoy10, jc.perez13, e.botero10, pa.arbelaez\}@uniandes.edu.co}
}

\maketitle

\begin{abstract}
We address the problem of segmenting an object given a natural language expression that describes it. Current techniques tackle this task by either (\textit{i}) directly or recursively merging linguistic and visual information in the channel dimension and then performing convolutions; or by (\textit{ii}) mapping the expression to a space in which it can be thought of as a filter, whose response is directly related to the presence of the object at a given spatial coordinate in the image, so that a convolution can be applied to look for the object. We propose a novel method that integrates these two insights in order to fully exploit the recursive nature of language. Additionally, during the upsampling process, we take advantage of the intermediate information generated when downsampling the image, so that detailed segmentations can be obtained. We compare our method against the state-of-the-art approaches in four standard datasets, in which it surpasses all previous methods in six of eight of the splits for this task. 

\end{abstract}

\section{Introduction}

\begin{figure}
	\centering
    \subfloat[Original image.]{{\includegraphics[width=0.25\columnwidth]{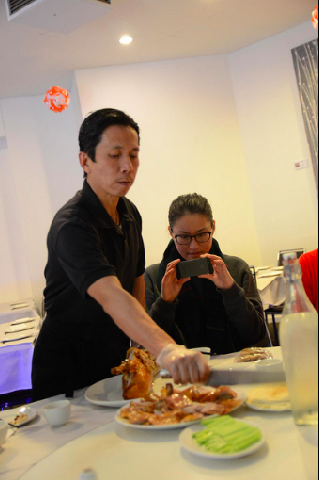} }}%
    \quad
    \subfloat[Output based on query \textit{guy}.]{{\includegraphics[width=0.25\columnwidth]{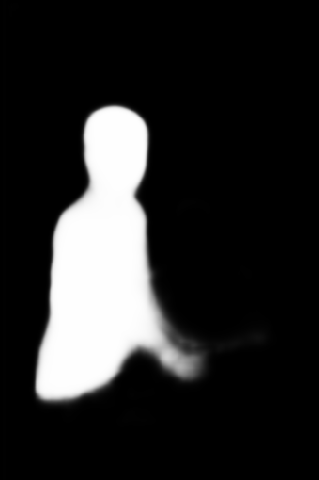} }}%
    \quad
    \subfloat[Output based on query \textit{girl}.]{{\includegraphics[width=0.25\columnwidth]{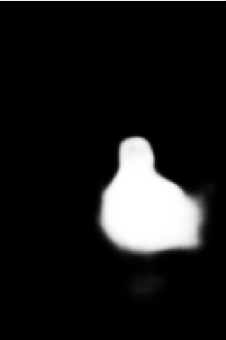} }}%
    \caption{\label{fig:query} Example of \textit{segmentation based on a natural language expression}. A single mask is the output, in which the only two labels are \textit{member of query} and \textit{background}. Here, we show the raw output of our system, which is the pixelwise probability of belonging to the referred object instance.}
\end{figure}

Consider the task of retrieving specific object instances from an image based  on natural language descriptions, as illustrated in Fig.~\ref{fig:query}. In contrast to traditional instance segmentation, in which the goal is to label all pixels belonging to instances in the image for a set of predefined semantic classes  \cite{DBLP:journals/corr/LinMBHPRDZ14}\cite{10.1007/978-3-319-10584-0_20}, segmenting instances described by a natural language expression is a task that humans are able to perform without specifically focusing on a limited set of categories: we simply associate a referring expression such as \say{Man on the right} with what we see, as shown in Fig.~\ref{fig:query}. To learn such an association is the main goal of this paper.

In this task, the main labels to be assigned are \textit{related to query} and \textit{background}. Thus, the set of possible segmentation masks has few constraints, as a mask can be anything one might observe in the image, in all the ways natural language allows an object to be referred to. An algorithm to tackle this problem must then make sense of the query and relate it to what it sees and recognizes in the image, to finally output an instance segmentation map. Therefore, attempting to naively use Convolutional Neural Networks (CNNs) for this task falls short, since such networks do not model sequential information by nature, as is required when processing natural language. Given that the cornerstone of this task is the proper \textit{combination} of information retrieved from multiple, dissimilar domains, we expect traditional architectures, like CNNs and Recurrent Neural Networks (RNNs), to be useful modules, but we still need to design an overall architecture that fully exploits their complementary nature. 



In this paper, we introduce a modular neural network architecture that divides the task into several sub-tasks, each handling a different type of information in a specific manner. Our approach is similar to \cite{hu2016segmentation}, \cite{liu2017segmentation} and \cite{li2017cvpr} in that we extract visual and natural language information in an independent manner by employing networks commonly used for these types of data, \textit{i.e.}, CNNs and RNNs, and then focus on processing this multi-domain information by means of another neural network, yielding an end-to-end trainable architecture. However, our method also introduces the usage of Simple Recurrent Units (SRUs) for efficient segmentation based on referring expressions, a Synthesis Module that processes the linguistic and visual information jointly, and an Upsampling Module that outputs highly detailed segmentation maps. 

\begin{figure*}[t]
\centering
\includegraphics[width=0.7\textwidth]{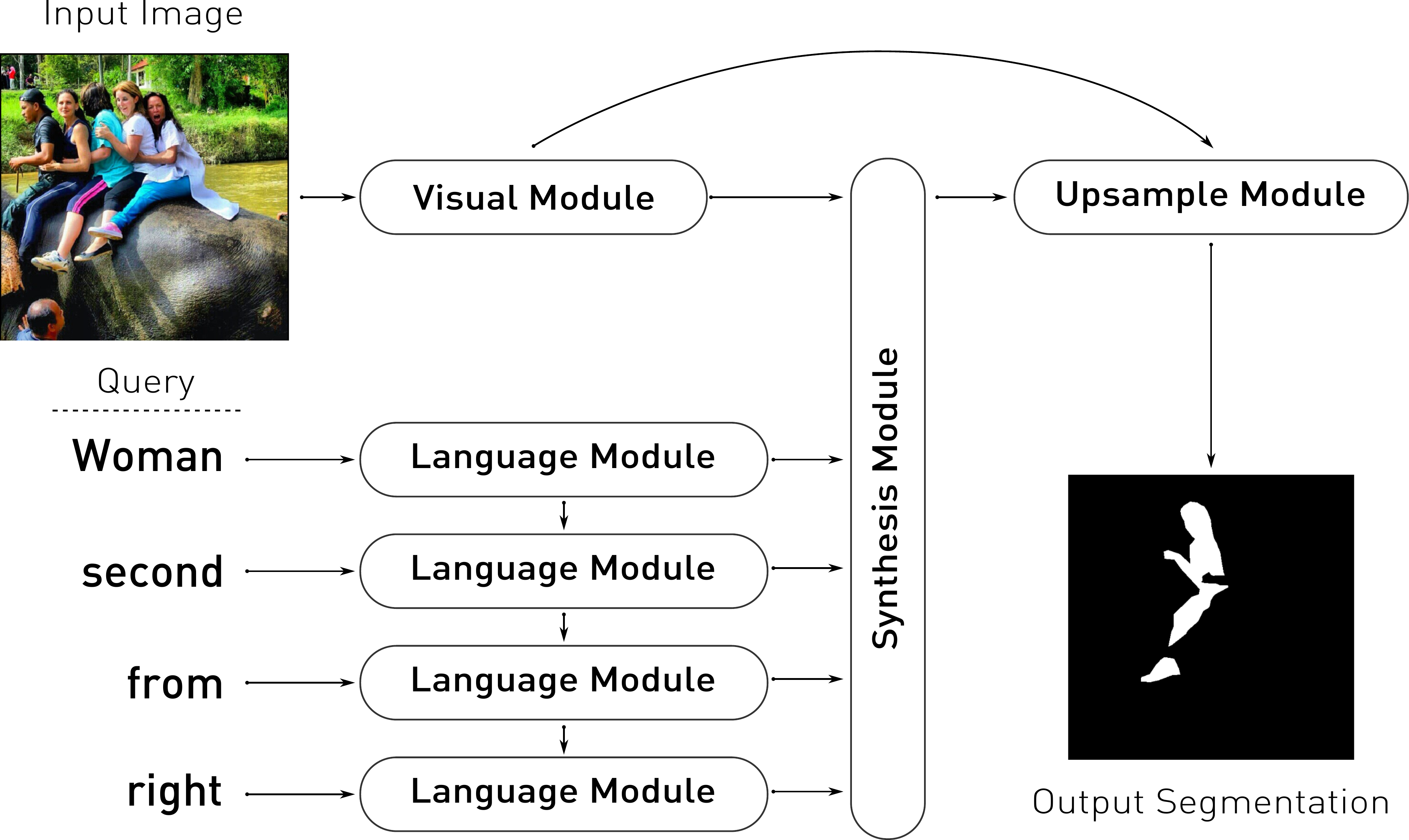}
\caption{Overview of our Dynamic Multimodal Network (DMN), involving four different modules: Visual Module (VM), Language Module (LM), Synthesis Module (SM), and Upsampling Module (UM).}
\label{fig:architecture}
\end{figure*}

Our network, which we refer to as Dynamic Multimodal Network (DMN), is composed of several modules, as depicted in Fig.~\ref{fig:architecture}: \textit{(i)} a Visual Module (VM) that produces an adequate representation of the image, \textit{(ii)} a Language Module (LM) that outputs an appropriate representation of the meaning of the query up to a given word, \textit{(iii)} a Synthesis Module (SM) that merges the information provided by the VM and LM at each time step and produces a single output for the whole expression and, finally, \textit{(iv)} an Upsampling Module (UM) that incrementally upsamples the output of the SM by using the feature maps produced by the VM. 
Our approach is a fully differentiable, end-to-end trainable neural network for segmentation based on natural language queries.
Our main contributions are the following:


\begin{itemize}
\item The use of Simple Recurrent Units (SRUs) \cite{DBLP:journals/corr/abs-1709-02755} as language \textit{and} multi-modal processors instead of standard LSTMs \cite{hochreiter1997long}. We empirically show that they are efficient while providing high performance for the task at hand.

\item A Synthesis Module that takes visual and linguistic information and merges them by generating \say{scores} for the referring expression in a visual space.

\item The Synthesis Module then takes this representation as well as additional features, and exploits the spatial and sequential nature of both types of information to produce a low resolution segmentation map. 

\item A high resolution upsampling module that takes advantage of visual features during the upsampling procedure in order to recover fine scale details.

\end{itemize}

We validate our method by performing experiments on all standard datasets, and show that DMN outperforms all the previous methods in various splits for instance segmentation based on referring expressions, and obtains state-of-the art results. Additionally, in order to ensure reproducibility, we provide full implementation of our method and training routines, written in PyTorch\footnote{\url{https://github.com/BCV-Uniandes/query-objseg}} \cite{paszke2017automatic}.

\section{Related Work}\label{subsect:related_work}
The intersection of Computer Vision (CV) and Natural Language Understanding (NLU) is an active area of research that includes multiple tasks such as object detection based on natural language expressions \cite{hu_natural_2015}\cite{guadarrama_understanding_2016}, image captioning \cite{hendricks_deep_2015}\cite{gan_semantic_2016}\cite{johnson_densecap:_2016}\cite{yang_review_2016} and visual question answering (VQA) \cite{goyal_making_2016}\cite{zhu_visual7w:_2015}\cite{krishna_visual_2016}\cite{agrawal_vqa:_2015}\cite{teney_graph-structured_2016}. Since visual and linguistic data have properties that make them fundamentally different, \textit{i.e.}, the former has spatial meaning and no sequentiality while the latter does not contemplate space but has a sequential nature, optimally processing both types of information is still an open question. Hence, each work in this sub-field has proposed a particular way of addressing each task.

The task studied in this paper is closest in nature to object detection based on natural language expressions, mirroring how semantic segmentation arose from object detection \cite{Everingham10}. Indeed, in \cite{hu2016segmentation}, object detection with NLU evolved into instance segmentation using referring expressions. We review the state-of-the-art on the task of segmentation based on natural language expressions \cite{hu2016segmentation}\cite{liu2017segmentation}\cite{li2017cvpr}, highlighting the main contributions in the fusion of multimodal information, and then compare them against our approach. 

\textbf{\emph{Segmentation from Natural Language Expressions \cite{hu2016segmentation}.}}\label{subsect:original} This work processes visual and natural language information through separate neural networks: a CNN extracts visual features from the image while an LSTM scans the query. Strided convolutions and pooling operations in the CNN downsample the feature maps to a low resolution output while producing large receptive fields for neurons in the final layers. Additionally, to explicitly model spatial information, relative coordinates are concatenated at each spatial location in the feature map obtained by the CNN. Merging of visual and natural language information is done by concatenating the LSTM's output to the visual feature map at each spatial location. Convolution layers with ReLU \cite{nair2010rectified} nonlinearities are applied for final classification. The loss is defined as the average over the per-pixel weighed logistic regression loss. Training has two stages: a low resolution stage, in which the ground truth mask is downsampled to have the same dimensions as the output, and a high resolution stage that trains a deconvolution layer to upsample the low resolution output to yield the final segmentation mask \cite{hu2016segmentation}. This seminal method does not fully exploit the sequential nature of language, as it does not make use of the learned word embeddings, it merges visual and linguistic information by concatenation, and it uses deconvolution layers for upsampling, which have been shown to introduce checkerboard artifacts in images \cite{odena2016deconvolution}.

\textbf{\emph{Recurrent Multimodal Interaction \cite{liu2017segmentation}.}} This paper argues that segmenting the image based only on a final, memorized representation of the  sentence does not fully take advantage of the sequential nature of language. Consequently, the paper proposes to perform segmentation multiple times in the pipeline. The method produces image features at every time step by generating a representation that involves visual, spatial and linguistic features. Such multimodal representation is obtained by concatenating the hidden state of the LSTM that processed the query at every spatial location of the visual representation. The segmentation mask is obtained by applying a multimodal LSTM (mLSTM) to the joint representation and then performing regular convolutions to combine the channels that were produced by the mLSTM. The mLSTM is defined as a convolutional LSTM that shares weights both across spatial location and time step, and is implemented as a $1\times1$ convolution that merges all these types of information. 
Bilinear upsampling is performed to the network's output at test time to produce a mask with the same dimensions of the ground-truth mask. This method reduces strides of convolutional layers and uses atrous convolution in the final layers of the CNN to compensate for the downsampling. Such modification reduces the upsampling process to bilinear interpolation, but can decrease the CNN's representation capacity while also increasing the number of computations that must be performed by the mLSTM.

\textbf{\emph{Tracking by Natural Language Specification \cite{li2017cvpr}.}} In this paper, the main task is object tracking in video sequences. A typical user interaction in tracking consists in providing the bounding box of the object of interest in the first frame. However, this type of interaction has the issue that, for the duration of the video, the appearance and location of objects may change, rendering the initial bounding box useless in some cases. The main idea is to provide an alternative to this approach, by noting that \textit{(i)} the semantic meaning of the object being tracked does not vary for the duration of the video as much as the appearance, and \textit{(ii)} this semantic meaning may be better defined by a linguistic expression. This approach is substantially different from \cite{liu2017segmentation} and \cite{hu2016segmentation}: visual and linguistic information is never merged \textit{per se}, but rather the linguistic information is mapped to a space in which it can be interpreted as having visual meaning.
The visual input is thus processed by a modified VGG \cite{vgg_paper} to yield a feature map. An LSTM scans the linguistic input, and a single layer perceptron is applied to the LSTM's last hidden state to generate a vector that can be interpreted as being a \textit{filter for a 2D convolution} that is to be performed on the feature map. 
The \textit{dynamic convolutional visual filter}, generated based on the expression, is computed to produce a strong response to the elements being referred to the expression, and a weak response to those \textit{not} being referred to. This response is interpreted as a \say{score} for the referring expression, so that a segmentation can be produced.
This method proposes a new paradigm for combining information from the visual and linguistic domains, but assumes a non linear combination of the last hidden state is sufficient for modeling a filter that responds to the query.

\textbf{\emph{Our approach.}} The approach of \cite{hu2016segmentation} merges multi-domain information by concatenation of linguistic information, subsequent $1\times1$ convolutions for segmentation and a deconvolution layer to perform upsampling. The method in \cite{liu2017segmentation} follows the same logic as \cite{hu2016segmentation} but introduces \textit{recursion} into the approach, exploiting the linguistic information further; however, the upsampling module is an interpolation that produces rather coarse results, to which the authors apply a post-processing DenseCRF, making the architecture not trainable end-to-end. Finally, \cite{li2017cvpr} has a different approach, in which linguistic information is never merged with feature maps, but is rather transformed so that it can detect the locations in the image to which the referring expression has strong response; nonetheless, like \cite{hu2016segmentation}, it does not fully exploit linguistic information in a sequential manner. Moreover, all these methods fail to utilize information acquired in the downsampling process in the upsampling process.

Our approach takes advantage of the previous insights, and consists of a modularized network that exploits both the possibility of segmentation based on combinations of multi-domain information, and the feasibility of producing filters that respond to objects being referred to by processing the linguistic information. Following the spirit of \cite{DBLP:journals/corr/RonnebergerFB15}\cite{DBLP:journals/corr/LongSD14}\cite{hypercolumns}, we use skip connections between the downsampling process and the upsampling module to output finely-defined segmentations. We employ the concatenation strategy of \cite{hu2016segmentation} but include richer visual and language features. Furthermore, we make use of dynamic filter computation, like \cite{li2017cvpr}, but in a sequential manner. Lastly, we introduce the use of a more efficient alternative to LSTMs in this domain, namely SRUs. We demonstrate empirically that SRUs can be used for modeling language \textit{and} multimodal information for this task, and that they can be up to $3\times$ faster than LSTMs, allowing us to train more expressive models.

\begin{figure*}[t]
\centering
\includegraphics[width=0.85\textwidth]{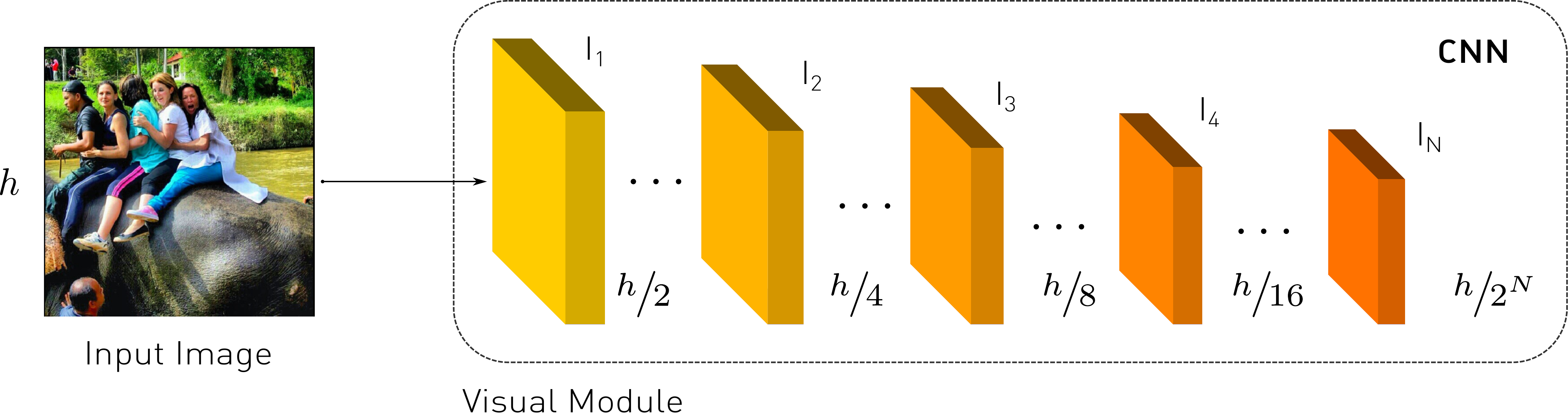}
\caption{The Visual Module outputs feature maps at $N$ different scales with the aim of using them in the segmentation process and in the upsampling.}
\label{fig:VM}
\end{figure*}

\section{Dynamic Multimodal Network}

\subsection{Overall Architecture}
Fig.~\ref{fig:architecture} illustrates our overall architecture. Given an input consisting of an image $I$, and a query composed of $T$ words, $\{w_{t}\}_{t=1}^{T}$, the Visual Module (VM) takes $I$ as input and produces feature maps at $N$ different scales: $\{I_{n}\}_{n=1}^{N}$. The Language Module (LM) processes $\{w_{t}\}_{t=1}^{T}$ and yields a set of features $\{r_{t}\}_{t=1}^{T}$ and a set of dynamic filters $\{\{f_{k,t}\}_{k=1}^{K}\}_{t=1}^{T}$. Given the VM's last output, $I_{N}$, $\{r_{t}\}_{t=1}^{T}$, and $\{\{f_{k,t}\}_{k=1}^{K}\}_{t=1}^{T}$, the Synthesis Module (SM) processes this information and produces a single feature map for the entire referring expression. This output, along with the feature maps given by the VM, is processed by the Upsampling Module (UM), that outputs a heatmap with a single channel, to which a sigmoid activation function is applied in order to produce the final prediction. 

\subsection{Visual Module (VM)}\label{sec:visual_module}
Fig.~\ref{fig:VM} depicts the Visual Module.  We extract deep visual features from the image using as backbone a Dual Path Network 92 (DPN92) \cite{DBLP:journals/corr/ChenLXJYF17}, which has shown competitive performance in various tasks, and is efficient in parameter usage. The VM can be written as a function returning a tuple:

\begin{equation}\label{eq:vm}
\{I_{n}\}_{n=1}^{N} = \text{VM}(I)
\end{equation}

Where $I$ is the original image, and $I_{n}, n \in \{1, \dotsc, N\}$ are the downsampled feature maps of dimensions equal to $\frac{1}{2^n}$ of the dimensions of $I$. In the experiments, we use $N=5$, which considers all convolutional layers in the visual encoder. Note that, since our architecture is fully convolutional, we are not restricted to a fixed image size.

\begin{figure*}[t]
\centering
\includegraphics[scale=0.17]{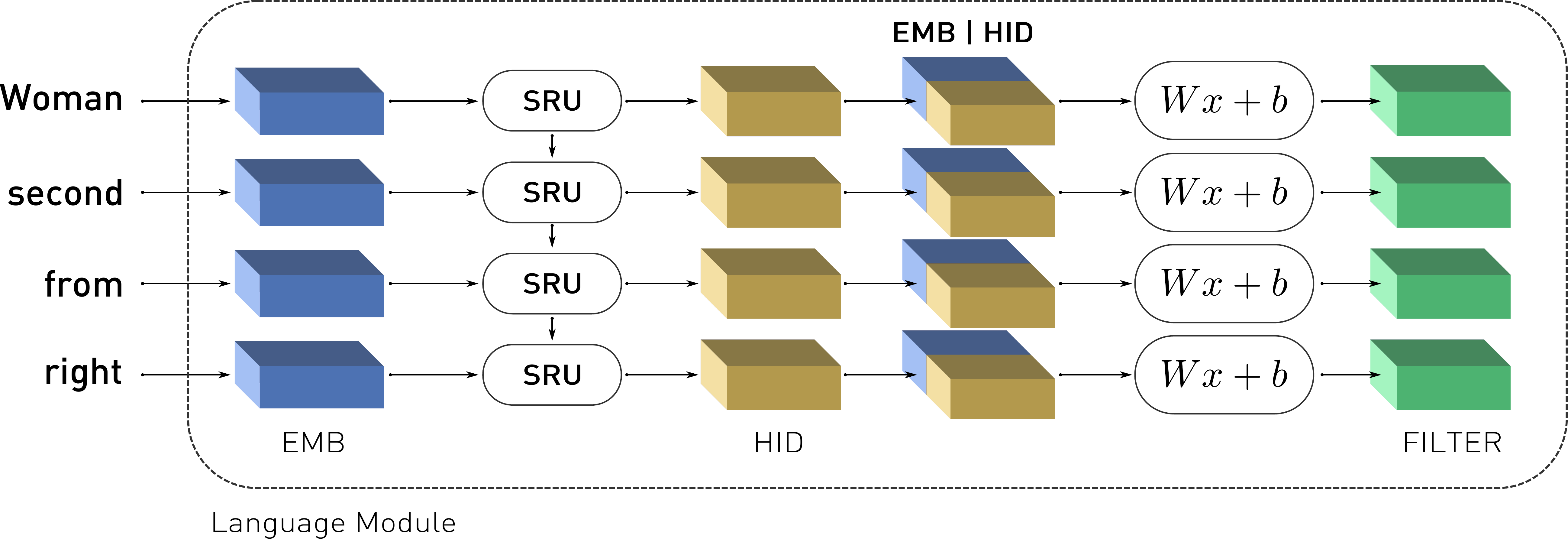}
\caption{The Language Module uses an SRU, instead of the traditional LSTM, to output enriched features of the query and dynamic filters based on such features.}
\label{fig:LM}
\end{figure*}

\subsection{Language Module (LM)}\label{sec:linguistic_module}
Fig.~\ref{fig:LM} shows a diagram of the Language Module. Given an expression consisting of $T$ words $\{w_{t}\}_{t=1}^{T}$, each word is represented by an embedding (WE), $e_{t} = WE(w_{t})$ ($EMB$ in Fig.~\ref{fig:LM}), and the sentence is scanned by an RNN to produce a hidden state $h_{t}$ for each word ($HID$ in Fig.~\ref{fig:LM}). Instead of using LSTMs as recurrent cells, we employ SRUs \cite{DBLP:journals/corr/abs-1709-02755}, which allow the LM to process the natural language queries more efficiently than when using LSTMs. The SRU is defined by:
\begin{align}
\tilde{x}_{t} &= W x_{t} \\
f_{t}' &= \sigma\left(W_{f} x_{t} + b_{f}\right) \\
r_{t} &= \sigma\left(W_{r} x_{t} + b_{r}\right) \\
c_{t} &= f_{t}' \odot c_{t-1} + (1 - f_{t}') \odot \tilde{x}_{t} \\
h_{t} &= r_{t} \odot g(c_{t}) + (1 - r_{t}) \odot x_{t}
\end{align}

Where $\odot$ is the element-wise multiplication. The function $g(\cdot)$ can be selected based on the task; here we choose $g(\cdot)$ to be the sigmoid function. For further details regarding the SRU definition and implementation, please refer to \cite{DBLP:journals/corr/abs-1709-02755}.


We concatenate the hidden state $h_{t}$ with the word embedding $e_{t}$ to produce the final language output: $r_{t} = [e_{t}, h_{t}]$. This procedure yields an enriched language representation of the concept of the sentence up to word $t$. Moreover, we compute a set of dynamic filters $f_{k,t}$ based on $r_{t}$, defined by:

\begin{equation}\label{eq:filters}
f_{k,t} = \sigma\left(W_{f_{k}}r_{t} + b_{f_{k}}\right), k = 1, ..., K
\end{equation}

thus, we define the LM formally as:
\begin{equation}
\left(\{r_{t}\}_{t=1}^{T}, \{\{f_{k,t}\}_{k=1}^{K}\}_{t=1}^{T}\right) = \text{LM}\left(\{w_{t}\}_{t=1}^{T}\right)
\end{equation}

\begin{figure*}[t]
\centering
\includegraphics[width=0.6\textwidth]{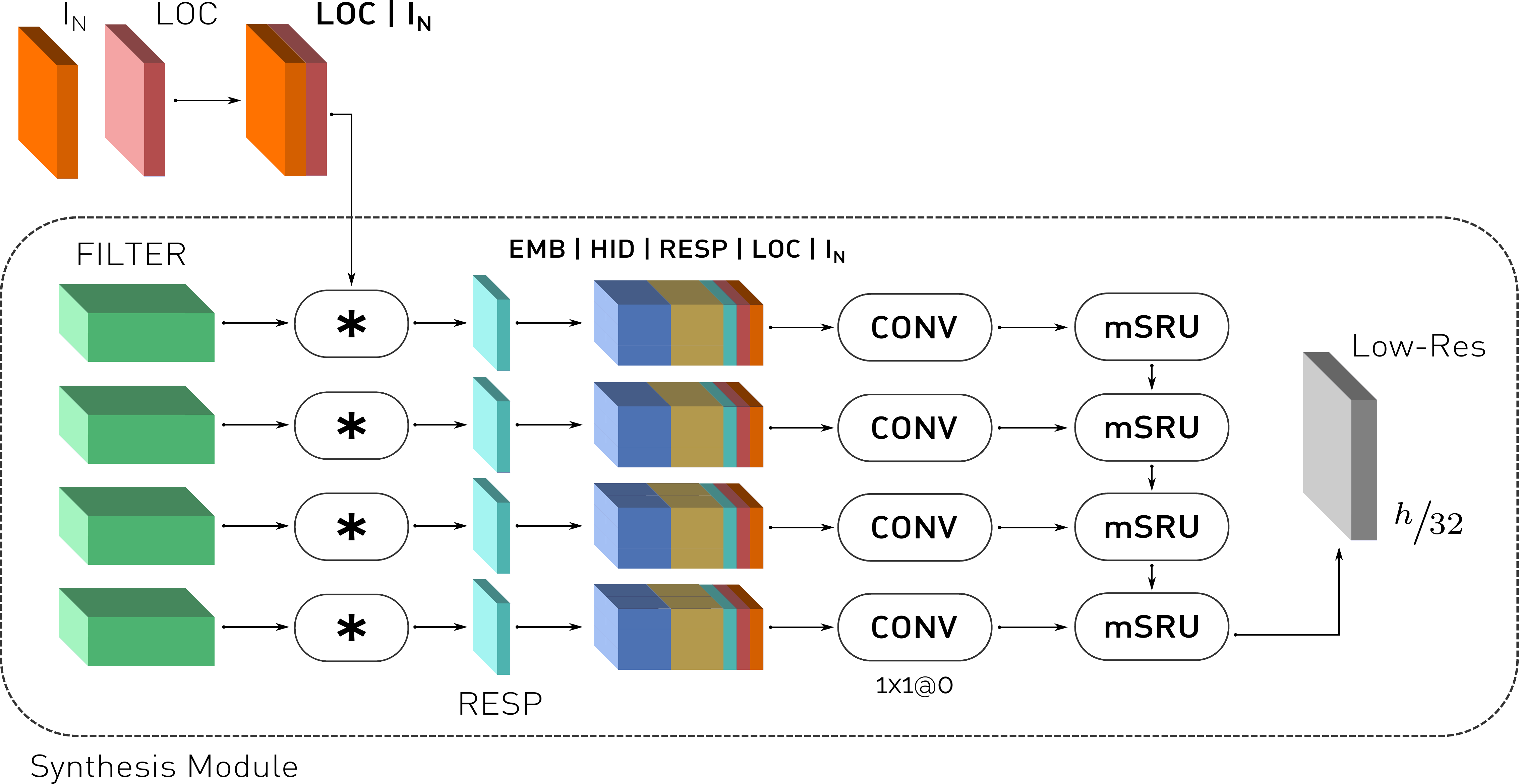}
\caption{\label{fig:RSM} The Synthesis Module takes into account the response to dynamic filters, language features, spatial coordinates representation, and visual features in a recurrent manner to output a single response map.}
\end{figure*}

\subsection{Synthesis Module (SM)}\label{sec:multimodal_module}
Fig.~\ref{fig:RSM} illustrates the Synthesis Module. The SM is the core of our architecture, as it is responsible for merging multimodal information. We first concatenate $I_{N}$ and a representation of the spatial coordinates ($LOC$ in Fig.~\ref{fig:RSM}), following the implementation of \cite{hu2016segmentation}, and convolve this result with each of the filters computed by the LM to generate a response map ($RESP$ in Fig.~\ref{fig:RSM}) consisting of $K$ channels: $F_{t} = \{f_{k,t} * I_{N}\}_{k=1}^{K}$. Next, we concatenate $I_{N}$, $LOC$, and $F_{t}$ along the channel dimension to obtain a representation $I'$, to which $r_{t}$ is concatenated \textit{at each spatial location}, as to have all the multimodal information in a single tensor. Finally, we apply a $1\times1$ convolutional layer that merges all the multimodal information, providing tan output corresponding to each time step $t$, denoted by $M_{t}$. Formally, $M_{t}$ is defined by:

\begin{equation}
M_{t} = \text{Conv}_{1\times1}([I_{N}, F_{t}, LOC, r_{t}])
\end{equation}

Next, in the pursuit of performing a recurrent operation that takes into account the sequentiality of the set and also the information of each of the channels in $M_{t}$, we propose the use of a multimodal SRU (mSRU), which we define as a $1\times1$ convolution, similar to \cite{liu2017segmentation} but using SRUs. We apply the mSRU to the whole set $\{M_{t}\}_{t=1}^{T}$, so that all the information in each $M_{t}$, including the sequentiality of the set, is used in the segmentation process. The final hidden states are gathered to produce a 3D tensor that is interpreted as a feature map. This tensor, which we denominate $R_{N}$, due to its size being $\frac{1}{2^N}$ of the image's original size, has the same dimensions as $M_{t}$ and has as many channels as there are entries in the hidden state of the mSRU. We define the SM as a function returning $R_{N}$:

\begin{equation}
R_{N} = \text{SM}\left(\{M_{t}\}_{t=1}^{T}\right) = \text{mSRU}\left(\{M_{t}\}_{t=1}^{T}\right),
\end{equation}

where $M_{t}$ is reshaped appropriately to make sense of the sequential nature of the information at each time step.

\subsection{Upsampling Module (UM)}\label{sec:upsampling_module}
\begin{figure*}[t]
\centering
\includegraphics[scale=0.14]{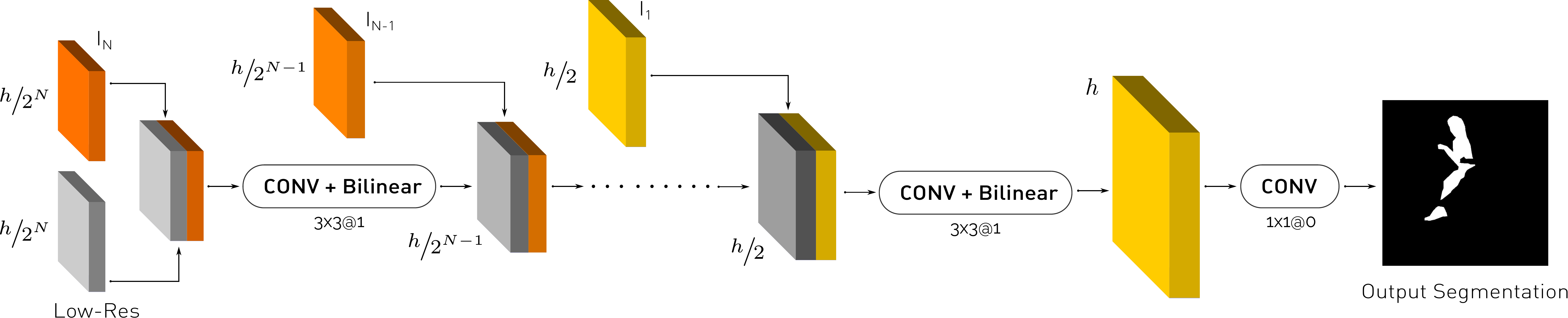}
	\caption{The Upsampling Module makes use of all the feature maps that were generated in the feature extraction process to provide more detailed segmentations.}
\label{fig:unet}
\end{figure*}

Finally, the Upsampling Module is shown in Fig.~\ref{fig:unet}. Inspired by skip connections \cite{DBLP:journals/corr/RonnebergerFB15}\cite{DBLP:journals/corr/LongSD14} \cite{DBLP:journals/corr/HuangLW16a}, we construct an upsampling architecture that takes into account the feature maps $\{I_{n}\}_{n=1}^{N}$ at all stages in order to recover fine-scale details. At each stage we concatenate $R_{n}$ with $I_{n}$, perform $3\times3$ convolution over this result, and then scale the size by a factor of \num{2} via bilinear interpolation to generate $R_{n-1}$. We apply this process $\log_{2}(N)$ times, to produce an output mask of the same size of the input $R_{1}$. We apply $1\times1$ convolution over $R_{1}$ to generate a single channel and, finally, a sigmoid layer to obtain scores between \num{0} and \num{1}. 

\section{Experimental Setup}
\subsection{Datasets}
We conduct experiments on the four standard datasets for this task: ReferIt, UNC, UNC+ \cite{KazemzadehOrdonezMattenBergEMNLP14}, and GRef \cite{DBLP:journals/corr/MaoHTCYM15}. UNC, UNC+ and GRef are based on MS COCO \cite{DBLP:journals/corr/LinMBHPRDZ14}. The type of objects that appear in the referring expressions, length of the expressions, and relative size of the referred objects are the main differences between the datasets. The high variability of those characteristics across the datasets evidences the challenge of constructing models for this task that are capable of generalization.

\textbf{ReferIt} \cite{KazemzadehOrdonezMattenBergEMNLP14} is a crowd-sourced database that contains images and referring expressions to objects in those images. Currently it has 130,525 expressions, referring to 96,654 distinct objects, in 19,894 photographs of natural scenes.


\textbf{UNC} \cite{DBLP:journals/corr/YuPYBB16}, was collected interactively in the ReferIt game, with images that were selected to contain two or more objects of the same object category \cite{DBLP:journals/corr/YuPYBB16}, which means that an expression making reference to a determined type of object will need to be further analysed to determine \textit{which} object the query is referring to, since ambiguity arises when only guided by semantic instance class cues. It consists of 142,209 referring expressions for 50,000 objects in 19,994 images.

\textbf{UNC+} \cite{DBLP:journals/corr/YuPYBB16}, is similar to UNC but has an additional restriction regarding words describing location: expressions must be based only on appearance rather than location. Such restriction implies that the expression will depend on the perspective of the scene and the semantic class of the object.

\textbf{GRef} \cite{DBLP:journals/corr/MaoHTCYM15}, was collected on Amazon's Mechanical Turk and contains 85,474 referring expressions for 54,822 objects in 26,711 images selected to contain between two and four objects of the same class, and thus, it presents similar challenges to those of UNC.



\subsection{Performance Metrics}
We use the standard metrics from the literature to allow for direct comparison with respect to the state-of-the-art. We perform experiments with the proposed method on the four standard datasets described above by training on the training set and evaluating the performance in each of the validation or test sets. We evaluate results by using two standard metrics: \textit{(i)} mean Intersection over Union (mIoU), defined as the total intersection area between the output and the Ground Truth (GT) mask, divided by the total union area between the output and the GT mask, added over all the images in the evaluation set, and \textit{(ii)} \textit{Precision}$@X$, or \textit{Pr}$@X$, ($X \in \{0.5, 0.6, 0.7, 0.8, 0.9\}$), defined as the percentage of images with IoU higher than $X$. We report mIoU in the validation and test splits of each dataset, when available, using optimal thresholds from the training or validation splits, respectively.

\subsection{Implementation Details}
All the models are defined and trained with DPN92 \cite{DBLP:journals/corr/ChenLXJYF17} as the backbone, which outputs \num{2688} channels in the last layer. We use $N = 5$ scales in the VM. We use the following hyperparameters, which we optimized on the UNC+ \textit{val} set: WE size of \num{1000}, \num{2}-layered SRU with hidden state size of \num{1000}, $K = 10$ filters, \num{1000} $1\times1$ convolution filters in the SM, \num{3}-layered mSRU's hidden state size of \num{1000}. The increased number of layers presented here in both the SRU and the mSRU, with respect to usual number of layers in LSTMs, are in response to the increased need of layers for an SRU to work as expected, according to \cite{DBLP:journals/corr/abs-1709-02755}. We train our method in two stages: at low resolution (\textit{i.e.}, without using the UM) and then finetune the UM to obtain high resolution segmentation maps. 

Training is done with Adam optimizer \cite{adam-opt} with an initial learning rate of \num{1e-5}, a scheduler that waits 2 epochs for loss stagnation to reduce the learning rate by a factor of \num{10}, and batch size of \num{1} image-query pair.

\section{Results}
\begin{table*}[b]
\centering
\caption{\label{tab:ablation-results} Precision$@X$ and mIoU for ablation study in the UNC testA split.}
\begin{tabular}{ c | p{1.2cm} p{1.2cm} p{1.2cm} p{1.2cm} p{1.2cm} | c }
\hline
Method 						& Pr$@0.5$			& Pr$@0.6$			& Pr$@0.7$			& Pr$@0.8$			& Pr$@0.9$		& mIoU			\\
\hline
Only VM 					& 15.26 			& 6.36 				& 2.96 				& 0.91 				& 0.14 			& 30.92 		\\
Only $h_{t}$ in LM and SM	& 65.38 			& \textbf{57.99} 	& \textbf{47.07}	& 27.38 			& 4.63			& 54.80 		\\
No skip connections in UM 	& 56.58 			& 42.77 			& 26.32 			& 9.22 				& 1.07 			& 49.26 		\\
No dynamic filters			& 57.53 			& 48.70 			& 38.27 			& 20.64 			& 3.00 			& 50.34 		\\
No concatenation of $r_{t}$	& 64.52 			& 56.69 			& 45.16				& 25.56 			& 4.38			& 54.69 		\\
\hline
DMN							& \textbf{65.83} 	& 57.82 			& 46.80 			& \textbf{27.64} 	& \textbf{5.12} & \textbf{54.83}\\
\hline
\end{tabular}
\end{table*}

\begin{table*}[t]
\centering
\caption{\label{tab:results} Comparison with the state-of-the-art in mIoU performance across the different datasets. Blank entries where authors do not report performance.}
\begin{tabular}{ c | c | c | c | c | c | c | c | c }
\hline
Method 	&\multicolumn{1}{c|}{Referit} 		& \multicolumn{1}{c|}{GRef} & \multicolumn{3}{c|}{UNC} 			& \multicolumn{3}{c}{UNC+} 		\\
							& test 			& val 						& val		& testA 	& testB 	& val 		& testA 	& testB 	\\
\hline
\cite{hu2016segmentation} 	& 48.03 		& 28.14						& - 		& -			& -			& -			& -			& - 		\\
\cite{Hu2016} 				& 49.91      	& 34.06						& - 		& -			& -			& -			& -			& - 		\\
\cite{li2017cvpr} 			& 54.30      	& - 						& - 		& -			& -			& -			& -			& - 		\\
\cite{liu2017segmentation}	&\bb{58.73}		& 34.52 					& 45.18		& 45.69 	&\bb{45.57} & 29.86		& 30.48 	& 29.50 	\\
\hline
DMN						& 52.81 		& \bb{36.76} 				&\bb{49.78}	& \bb{54.83}& 45.13  	&\bb{38.88}	&\bb{44.22}	& \bb{32.29} \\
\hline
\end{tabular}
\end{table*}
\subsection{Control Experiments}

We assess the relative importance of our modules in the final result by performing ablation experiments. The control experiments were trained until convergence on the UNC dataset. Accordingly, we compare them to a version of our full method trained for a similar amount of time. Table~\ref{tab:ablation-results} presents the results. 

The \say{Only VM} experiment in row 1 consists on training only the VM and upsampling the low resolution output with bilinear interpolation, without using the query. At test time, the VM processes the image and the resulting segmentation map is upsampled using the UM and compared to the GT mask. Results show how this method performs poorly in comparison to our full approach, which confirms our hypothesis that naively using a CNN falls short for the task addressed in this paper. However, it is interesting that performance is rather high for a method that does not use linguistic information at all. This result reveals that many of the objects annotated in this dataset are salient, and so the network is able to learn to segment salient objects without help from the query. 

The experiment in row 2 consists of defining $r_{t} = h_{t}$, instead of using the concatenation of $h_{t}$ and $e_{t}$, which affects both the LM (when computing the dynamic filters) and the SM. Results show that using the learned embeddings provides a small gain in the full method, particularly for stricter overlap thresholds.

Next, in row 3 we assess the importance of skip connections in the UM, which is a measure of the usefulness of features extracted in the downsampling process for the upsampling module. The large drop in performance with respect to the full method shows that the skip connections allow the network to exploit finer details that are otherwise lost, showing how the upsampling strategy can benefit from performing convolutions followed by bilinear interpolations instead of deconvolutions, as done in \cite{hu2016segmentation}.



We next study the effects of removing features from $M_t$. In rows 4 and 5 we remove the set of responses to the dynamic filters $F$, as well as the concatenation of $r_t$ in the SM, respectively. We observe that the dynamic filters generate useful scores for the natural language queries in the visual space, and that reusing features from the LM in the SM does not help the network significantly.



Our results show that the key components of our network have significant impact in overall performance. High performance is not achieved by either using only linguistic information ($r_{t}$) or the response to filters ($F$): both must be properly combined. Additionally, the UM allows the network to properly exploit features from the downsampling stage and perform detailed segmentation. 


\subsection{Comparison with the State-of-the-Art}

Next, we proceed to compare our full method with the state-of-the-art, for which we evaluate on all the datasets described above. Table~\ref{tab:results} compares the mIoU of our method with the state-of-the-art in this task \cite{hu2016segmentation}\cite{liu2017segmentation}\cite{li2017cvpr}. The results show that our method outperforms all other methods in six out of eight splits of the datasets. By including enriched linguistic features at several stages of the process, and by combining them in different ways, our network learns appropriate associations between queries and the instances they refer to. 

Interestingly, the performance gain in the \textit{testB} splits of UNC and UNC+ is not as large as in \textit{testA}. One possible reason for the smaller performance gain across splits is their difference: visual inspection of results shows how \textit{testA} splits are biased towards queries related to segmenting persons. The \textit{testB} splits, however, contain more varied queries and objects, which is why the increase in mIoU is not as marked. This behavior can also be observed for the method proposed by \cite{liu2017segmentation}, as shown in the second-to-last line of Table \ref{tab:results}. 

\begin{figure*}[t]
    \centering
    \subfloat[Training time.]{{\includegraphics[width=0.45\textwidth]{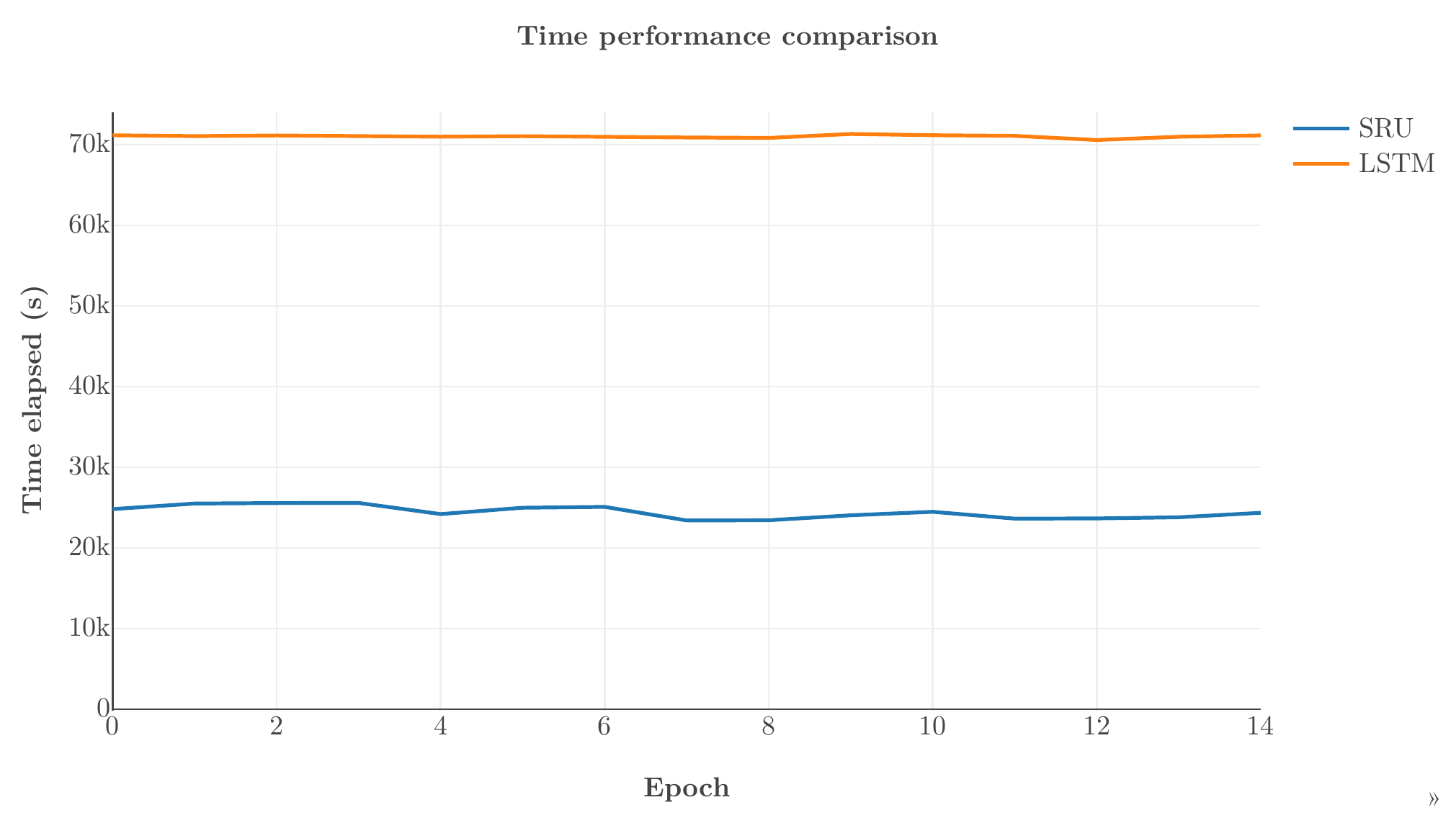} }}%
    \quad
    \subfloat[Performance (mIoU).]{{\includegraphics[width=0.45\textwidth]{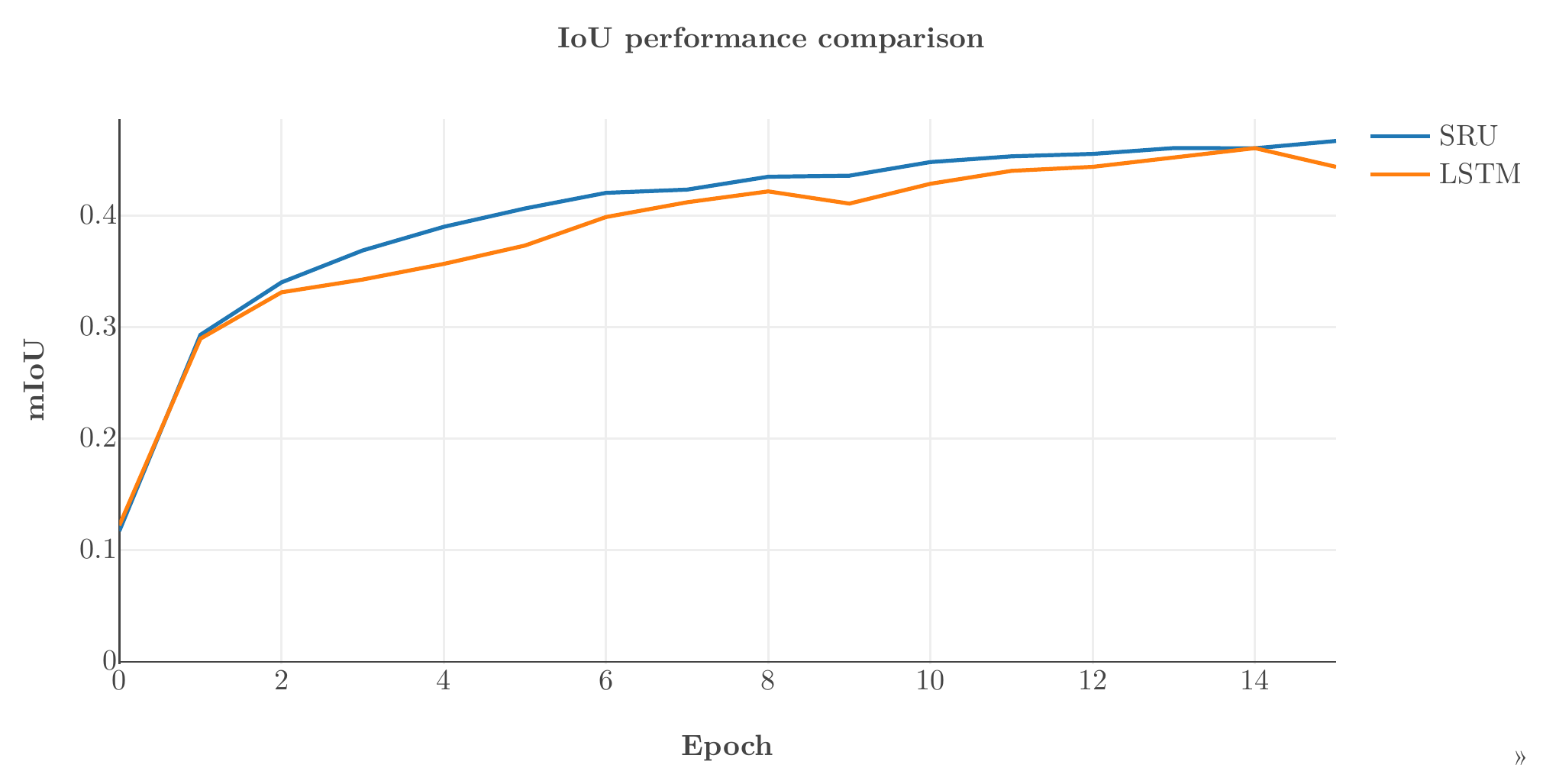} }}%
    \caption{SRUs \textit{vs.} LSTMs comparison on UNC \textit{testA} (at low resolution).}
    \label{Fig:SRU_vs_LSTM}
\end{figure*}

\subsection{Efficiency Comparison: SRU \textit{vs.} LSTM}
In order to assess the efficiency and the performance of SRUs when compared to the more commonly used LSTMs, both as language \textit{and} multi-modal processors, we conduct an experiment in which we replaced the SRUs with LSTMs in our final system, both in the LM and the SM, we trained on the UNC dataset, and we measured performance on the \textit{testA} split. In terms of model complexity, when using SRUs, the LM and the SM have 9M and 10M trainable parameters, respectively. When switching to LSTMs, the number of parameters increases to 24M and 24.2M, respectively, multiplying training time by a factor of three, as shown in Fig.~\ref{Fig:SRU_vs_LSTM}\textit{a}. Regarding accuracy, Fig.~\ref{Fig:SRU_vs_LSTM}\textit{b}. shows that both  systems perform similarly, with a small advantage for SRUs. Therefore, when compared to LSTMs, SRUs allow us to design architectures that are more compact, train significantly faster, and generalize better.



\begin{figure*}[t]
\begin{centering}
	\subfloat[\textit{yellow shirt}]{{\includegraphics[width=0.145\columnwidth]{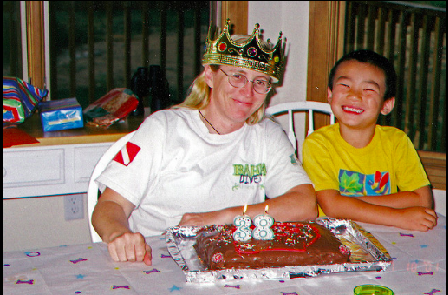} }{\includegraphics[width=0.148\columnwidth]{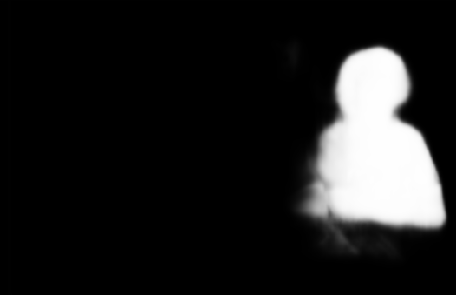} }{\includegraphics[width=0.145\columnwidth]{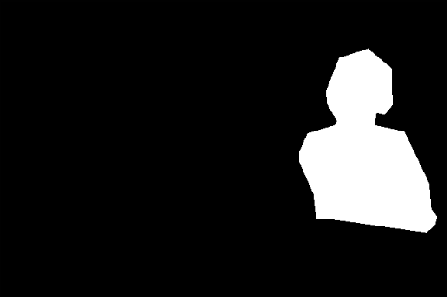} }}%
    \quad
    \subfloat[\textit{man alone on the right}]{{\includegraphics[width=0.133\columnwidth]{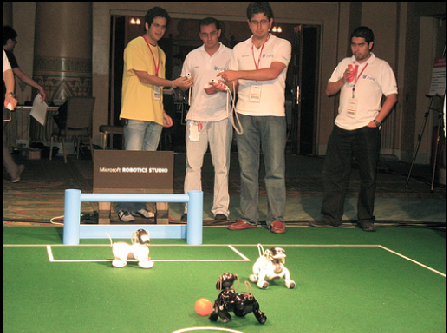} }{\includegraphics[width=0.133\columnwidth]{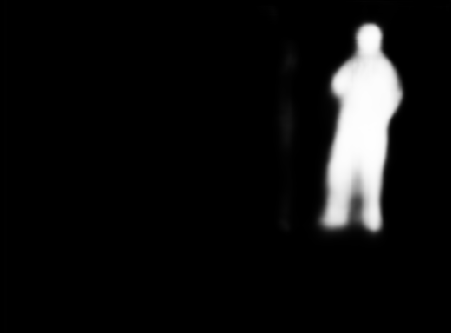} }{\includegraphics[width=0.133\columnwidth]{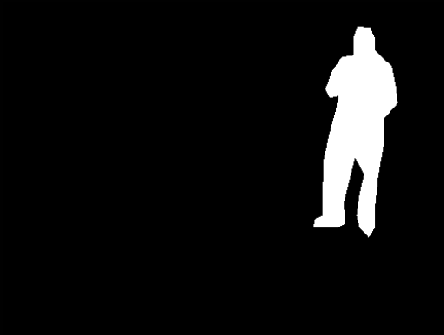} }}%
    \quad
    \subfloat[\textit{batter}]{{\includegraphics[width=0.145\columnwidth]{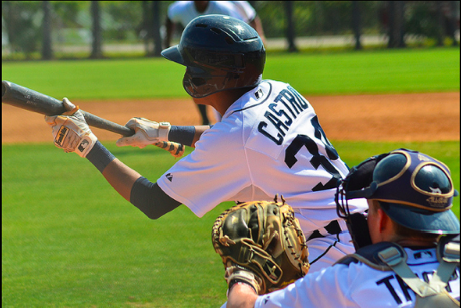} }{\includegraphics[width=0.145\columnwidth]{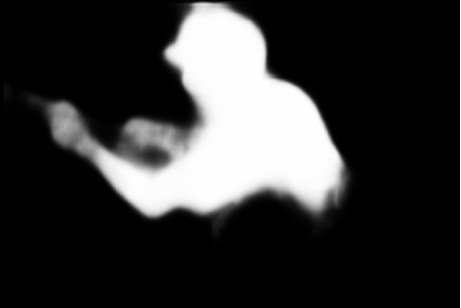} }{\includegraphics[width=0.145\columnwidth]{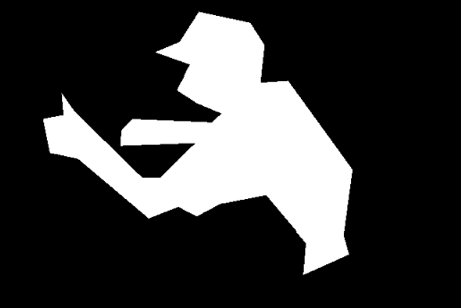} }}%
    \quad
    \subfloat[\textit{catcher}]{{\includegraphics[width=0.125\columnwidth]{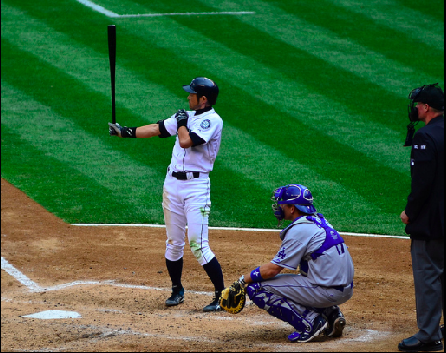} }{\includegraphics[width=0.125\columnwidth]{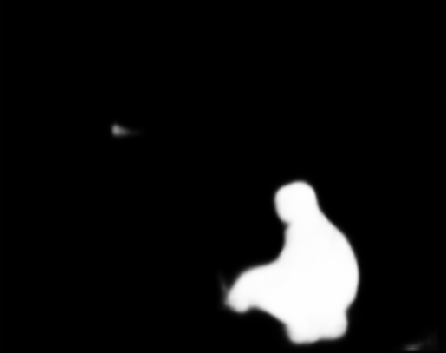} }{\includegraphics[width=0.125\columnwidth]{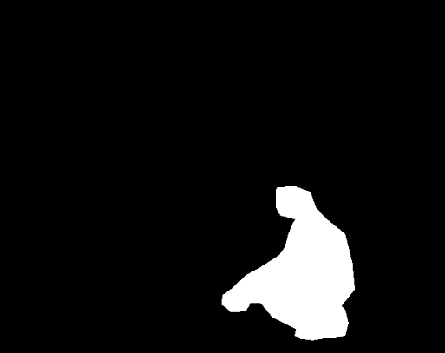} }}%
    
\end{centering}
    \caption{\label{fig:positive_examples} Qualitative examples of the output of the network. From left to right in each subfigure: original image, heatmap produced by our method, and ground-truth mask. Each caption is the query that produced the output.}
\end{figure*}

\subsection{Qualitative Results}
Fig.~\ref{fig:positive_examples} shows qualitative results in which the network performed well. These examples demonstrate DMN's flexibility for segmenting based on different information about a particular class or instance: attributes, location or role. We emphasize that understanding a role is not trivial, as it is related to the object's context and appearance. Additionally, a semantic difficulty that our network seems to overcome is that the role coexists with the object's class: an instance can be \say{batter} and be \say{person}.
Notice in Fig.~\ref{fig:positive_examples} that thanks to the upsampling module, our network segments fine details such as legs, heads and hands. In Fig.~\ref{fig:positive_examples}\textit{a} the query refers to the kid by one of his \textit{attributes}: the color of his shirt; in Fig.~\ref{fig:positive_examples}\textit{b} the man is defined by his \textit{location} and the fact that he is alone (although that could be dropped, as there is no ambiguity); in Figs.~\ref{fig:positive_examples}\textit{c} and \textit{d} the reference is based on the person's \textit{role}. 

\begin{figure*}[h]
\begin{centering}
    \subfloat[\textit{person on left}]{{\includegraphics[width=0.14\columnwidth]{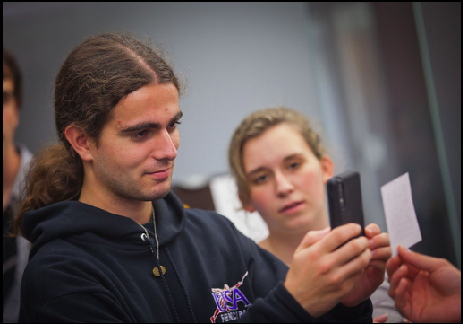} }{\includegraphics[width=0.14\columnwidth]{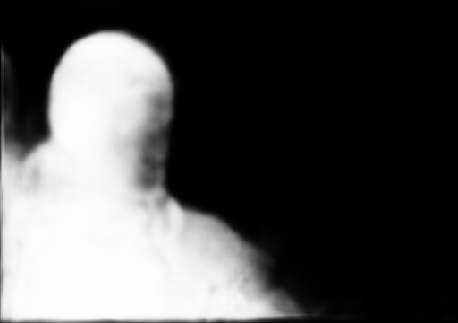} }{\includegraphics[width=0.14\columnwidth]{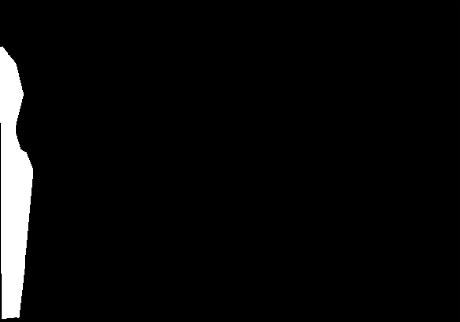} }}%
    \quad
    \subfloat[\textit{person sitting on the right with a hat that has a white stripe}]{{\includegraphics[width=0.125\columnwidth]{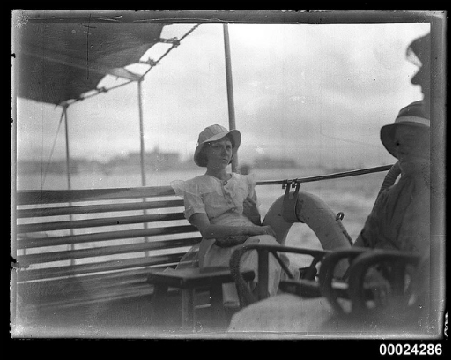} }{\includegraphics[width=0.125\columnwidth]{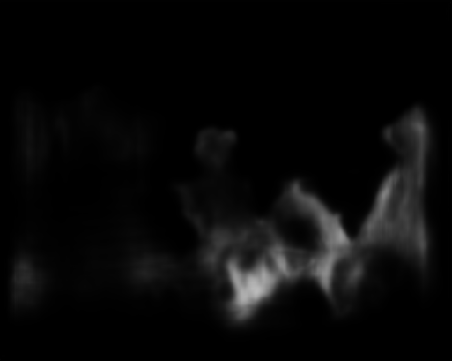} }{\includegraphics[width=0.125\columnwidth]{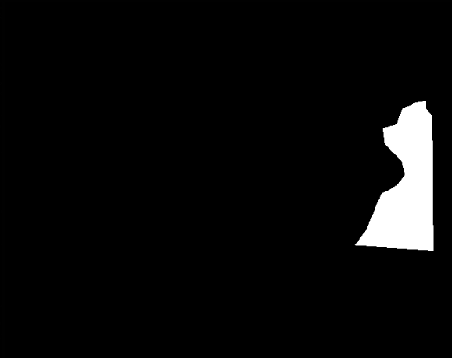} }}%
    \quad
    \subfloat[\textit{guy in gray shirt standing}]{{\includegraphics[width=0.114\columnwidth]{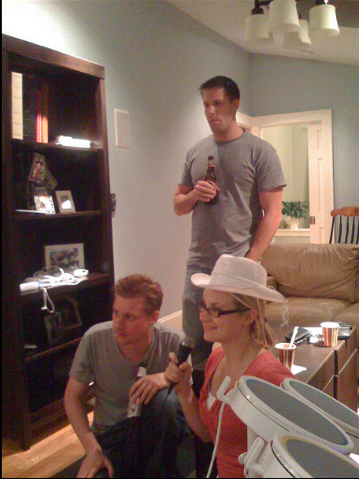} }{\includegraphics[width=0.114\columnwidth]{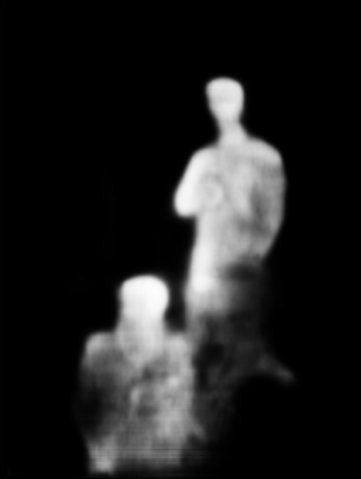} }{\includegraphics[width=0.114\columnwidth]{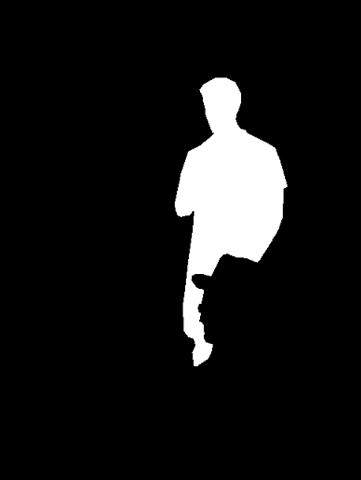} }}%
    \quad
    \subfloat[\textit{hand on remote}]{{\includegraphics[width=0.092\columnwidth]{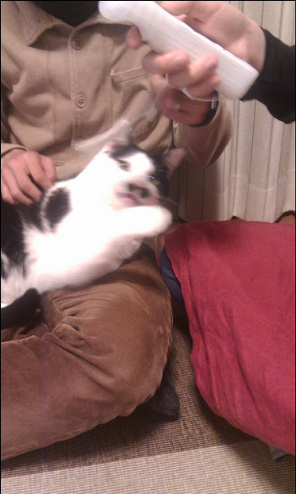} }{\includegraphics[width=0.092\columnwidth]{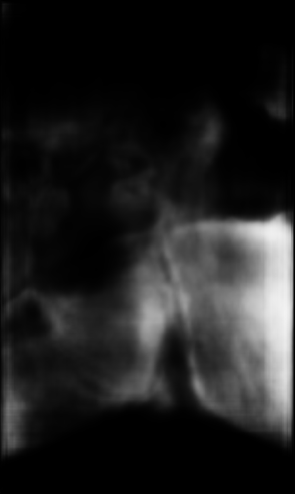} }{\includegraphics[width=0.092\columnwidth]{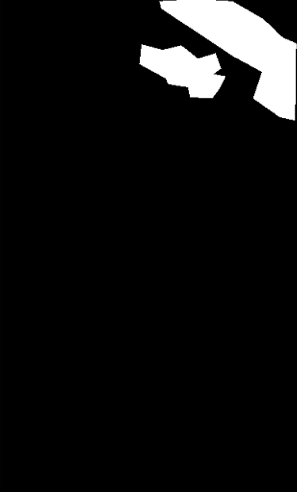} }}%

\end{centering}
    \caption{\label{fig:negative_examples} Negative examples of the output of the network. From left to right in each Subfigure: original image, heatmap produced by our method, and ground-truth mask. Each caption is the query that produced the output.}
\end{figure*}
Typical failure cases are depicted in Fig.~\ref{fig:negative_examples}. In Fig.~\ref{fig:negative_examples}\textit{a} the network segments (arguably) the incorrect person, since the correct segmentation was the person at the border of the image whose face is partially shown. Several failure cases we found had exactly the same issue: ambiguity in the expression that could confuse even a human. Fig.~\ref{fig:negative_examples}\textit{b} shows an example of strong failure, in which a weak segmentation is produced. The model appears to have only focused on the word \say{right}. We attribute this failure to the network's inability to make sense of such a relatively long sentence, which, while unambiguously defining an object, is a confusing way of referring to it. 
Fig.~\ref{fig:negative_examples}\textit{c} is an interesting example of the network's confusion. While the woman is not segmented, two subjects that share several attributes (\textit{guy}, \textit{gray} and \textit{shirt}) are confused and are \textit{both} segmented. However, the network does not manage to use the word \say{standing} to resolve the ambiguity. Finally, in Fig.~\ref{fig:negative_examples}\textit{d} a failure is observed, where nothing related to the query is segmented. The mask that is produced only reflects a weak attempt of segmenting the red object, while ignoring the upper part of the image, in which both the hand \textit{and} the remote were present.



\section{Conclusions}
We propose Dynamic Multimodal Network, a novel approach for segmentation of instances based on natural language expressions. DMN integrates insights from previous works into a modularized network, in which each module has the responsibility of handling information from a specific domain. Our Synthesis Module combines the outputs from previous modules and handles this multi-modal information to produce features that can be used by the Upsampling Module. Thanks to the incremental use of feature maps obtained in the encoding part of the network, the Upsampling Module delivers great detail in the final segmentations. Our method outperforms the state-of-the-art methods in six of the eight standard dataset splits for this task. 

\textbf{Acknowledgments}\\
The authors gratefully thank NVIDIA for donating the GPUs used in this work.

\clearpage

\bibliographystyle{splncs}
\bibliography{egbib}
\end{document}